\documentclass[runningheads]{llncs}
\usepackage[T1]{fontenc}
\usepackage{graphicx}
\usepackage{booktabs}
\usepackage[misc]{ifsym}
\newcommand{\corr}{(\Letter)}

\usepackage{url}
\usepackage{hyphenat}
\usepackage{xurl} 
\usepackage[table]{xcolor}
\usepackage{arydshln}

\usepackage{amsmath,amssymb}
\usepackage{esvect} 
\usepackage[final]{microtype}
\usepackage{pifont}
\usepackage{multirow}
\usepackage{array}
\usepackage{colortbl}

\usepackage{hyperref}

\providecommand{\sd}[1]{{$\pm$#1}}

\begin{document}

\title{SpikF-GO: Spiking Fourier Graph Operators for Multivariate Time Series Forecasting}

\titlerunning{SpikF-GO: Spiking Fourier Graph Operators}

\author{Jafar Bakhshaliyev\corr \and Niels Landwehr}
\authorrunning{J. Bakhshaliyev and N. Landwehr}
\institute{Data Science Group, University of Hildesheim, Hildesheim, Germany\\
\email{bakhshaliyevj@uni-hildesheim.de, landwehr@uni-hildesheim.de}}

\maketitle   

\begingroup
  \renewcommand\thefootnote{}%
  \footnotetext{This paper has been accepted for presentation at ECML--PKDD 2026.}%
\endgroup

\begin{abstract}
Spiking Neural Networks (SNNs) have emerged as an energy-efficient alternative to conventional neural networks, demonstrating strong performance in computer vision and robotics. More recently, SNNs have been applied to time series forecasting (TSF), with methods exploring spiking temporal backbones, spike-compatible positional encodings, Fourier-domain processing, and redesigned neuron dynamics. However, existing SNN forecasting approaches process variables independently, lacking explicit mechanisms for modeling inter-variable dependencies. This is a critical limitation in multivariate settings, where cross-variable correlations carry substantial predictive information. We propose Spiking Fourier Graph Operators (SpikF-GO), which addresses this gap by combining a hypervariate graph formulation in which every scalar observation becomes a graph node with spike-driven spectral processing. SpikF-GO introduces a Hard Concrete frequency gate for learnable sparse frequency selection and a Complex LIF gate that applies independent spiking neurons to real and imaginary Fourier components, preserving binary, event-driven computation throughout the spectral domain. We further present a variant incorporating Central Pattern Generator-based positional encodings for stronger long-range temporal modeling. Evaluated on eight benchmarks under a unified experimental protocol, SpikF-GO achieves the best average rank among all SNN methods and outperforms its ANN counterpart, FourierGNN, at reduced energy cost. SpikF-GO maintains competitive accuracy even at substantially smaller embedding dimensions, thereby achieving significant energy reductions. To our knowledge, this is among the first works to bring graph-based multivariate modeling into the spiking domain for TSF and the first to provide a unified comparison across SNN forecasting architectures under a common experimental protocol.
\end{abstract}

\section{Introduction}

Spiking Neural Networks (SNNs), considered the third generation of neural networks~\cite{MAASS19971659}, have attracted significant attention for their energy efficiency, sparsity, and event-driven processing~\cite{eshraghian2023trainingspikingneuralnetworks,osti_1881146}. Unlike Artificial Neural Networks (ANNs) that operate on continuous-valued activations, SNNs communicate via discrete spike events, mimicking how biological neurons transmit information through action potentials~\cite{wu2025spikf,MAASS19971659,eshraghian2023trainingspikingneuralnetworks}. This event-driven paradigm enables SNNs to process information only when necessary, offering substantial computational savings over ANNs whose continuous operations create significant challenges for resource-constrained and edge-deployment environments~\cite{Roy2019,osti_1881146}.

Leveraging these efficiency advantages, SNNs have achieved significant progress across machine learning domains, particularly in computer vision. Spiking Transformer architectures have been applied to image classification~\cite{Yao_2025,zhou2024qkformerhierarchicalspikingtransformer,lee2025spikingtransformerspatialtemporalattention}, object detection~\cite{luo2025integervaluedtrainingspikedriveninference}, and semantic segmentation~\cite{lei2024spike2formerefficientspikingtransformer}, in several cases matching or surpassing ANN counterparts at a fraction of the energy cost. Beyond static inputs, the inherent temporal dynamics of spiking neurons make SNNs naturally suited for sequential and time-varying data, including time series forecasting~\cite{feng2025tsliftemporalsegmentspiking}.

Recent work on SNN-based time series forecasting has developed spiking counterparts of major temporal backbones. Lv~et~al.~\cite{lv2024efficienteffectivetimeseriesforecasting} introduced spiking TCN, RNN, and Transformer models with competitive accuracy and lower energy consumption, while Central Pattern Generator (CPG)-based positional encodings were proposed to address the permutation-invariance issue of spiking self-attention~\cite{lv2024advancingspikingneuralnetworks}. The Temporal Segment LIF (TS-LIF) neuron~\cite{feng2025tsliftemporalsegmentspiking} further improves multi-timescale integration through a dual-compartment design that separates low- and high-frequency processing. More recently, SpikF~\cite{wu2025spikf} showed that Fourier-domain processing is well suited to SNN forecasting by using a Spiking Fast Fourier Transform (S-FFT) for frequency selection, avoiding permutation invariance while achieving strong energy--accuracy trade-offs. In parallel, on the ANN side, FourierGNN~\cite{yi2023fouriergnnrethinkingmultivariatetime}, which is considered one of the state-of-the-art GNN-based forecasting architectures (see Section~\ref{sec:related_gnn}), proposed a hypervariate graph formulation that unifies spatial and temporal modeling through Fourier Graph Operators (FGOs), achieving log-linear complexity without separate graph and temporal modules. Despite this progress, existing SNN forecasting methods still process each variable independently and lack explicit modeling of inter-variable dependencies. In multivariate time series, cross-variable correlations often carry substantial predictive information, such as correlated sensor readings in traffic networks or co-moving patterns in energy grids, and ignoring these relationships leaves significant predictive signal unexploited \cite{chen2023tsmixerallmlparchitecturetime}. Furthermore, prior SNN forecasting studies lack a unified experimental comparison across architectures under consistent settings, making it difficult to assess the true relative strengths of different approaches.

In this paper, we propose \textbf{Spik}ing \textbf{F}ourier \textbf{G}raph \textbf{O}perators \textbf{(SpikF-GO)}, which addresses both limitations by combining the unified hypervariate graph formulation of FourierGNN~\cite{yi2023fouriergnnrethinkingmultivariatetime} with spike-driven Fourier-domain processing. Rather than processing variables or time steps through separate modules, SpikF-GO treats each scalar observation in the input window as a node in a hypervariate graph, enabling the model to learn joint spatiotemporal dependencies through spectral graph convolution in a single unified structure. This formulation addresses the lack of explicit inter-variable modeling in prior SNN forecasting methods such as TS-LIF~\cite{feng2025tsliftemporalsegmentspiking} and SpikF~\cite{wu2025spikf}. To maintain energy efficiency within this graph-based framework, SpikF-GO employs a Hard Concrete frequency gate~\cite{louizos2018learningsparseneuralnetworks} for learnable sparse frequency selection and a Complex LIF gate that applies independent spiking neurons to the real and imaginary components of the Fourier spectrum, preserving binary, event-driven computation throughout the spectral domain. In addition, we evaluate a variant, \textbf{SpikF-GO w/ CPG}, which injects CPG positional encodings~\cite{lv2024advancingspikingneuralnetworks} prior to spectral mixing to strengthen long-range temporal modeling. In summary, the main contributions are as follows:

\begin{itemize}
    \item We propose \textbf{SpikF-GO} (\textbf{Spik}ing \textbf{F}ourier \textbf{G}raph \textbf{O}perators), a spiking architecture for multivariate time series forecasting that combines a hypervariate graph formulation with spike-driven Fourier graph processing to model intra-series temporal dependencies, inter-series dependencies, and time-varying cross-variable interactions. We further introduce \textbf{SpikF-GO w/ CPG} to improve long-range temporal modeling.

    \item On eight benchmarks, \textbf{SpikF-GO w/ CPG} achieves the best average rank on both $R^2$ (\textbf{2.4}) and MAE (\textbf{2.3}), while \textbf{SpikF-GO} achieves the second-best average rank on $R^2$ (\textbf{2.8}), outperforming prior SNN baselines and surpassing the ANN baseline FourierGNN.
    \item We provide extensive ablations and theoretical energy analysis, showing that SpikF-GO achieves \textbf{1.89$\times$} lower theoretical energy than FourierGNN, and up to \textbf{7.86$\times$} lower with a compact embedding size of \textbf{$E{=}8$}.
\end{itemize}

The code is publicly available at \url{https://github.com/jafarbakhshaliyev/SpikF-GO}. We further release \textsc{SpikingTSF}\footnote{\url{https://github.com/spikora/SpikingTSF}}, an open-source library for SNN-based time series forecasting that integrates several SNN forecasting architectures under a unified training and evaluation protocol.

\section{Related Work}

\subsection{GNNs and Frequency-Based Models for Multivariate TSF}\label{sec:related_gnn}

Multivariate time series forecasting (TSF) has traditionally been approached with temporal architectures such as recurrent networks, convolutional models, and more recently Transformers~\cite{salinas2019deeparprobabilisticforecastingautoregressive,bai2018empiricalevaluationgenericconvolutional,wu2022autoformerdecompositiontransformersautocorrelation,zhou2021informerefficienttransformerlong}.
In parallel, graph-based forecasting methods leverage the observation that many multivariate systems exhibit structured inter-series dependencies, and therefore model both temporal dynamics and cross-variable interactions using Graph Neural Networks (GNNs)~\cite{yi2023fouriergnnrethinkingmultivariatetime}.

Early spatio-temporal GNN models such as STGCN and TAMP-S2GCNets rely on a predefined graph to encode spatial correlations \cite{Yu_2018,Chen2022TAMPS2GCNetsCT}. However, in many real-world forecasting problems the underlying dependency graph is unknown or time-varying. To address this, later approaches learn inter-series relations directly from data, including StemGNN~\cite{cao2021spectraltemporalgraphneural}, MTGNN~\cite{wu2020connectingdotsmultivariatetime}, and AGCRN~\cite{NEURIPS2020_ce1aad92}.
Despite their effectiveness, many of these methods still adopt a two-stream design---a graph module (e.g., GCN/GAT) for cross-series interactions and a temporal backbone (e.g., RNN/GRU/LSTM) for temporal dependencies---which increases architectural complexity and can complicate optimization~\cite{yi2023fouriergnnrethinkingmultivariatetime}.

FourierGNN~\cite{yi2023fouriergnnrethinkingmultivariatetime} addresses this limitation by proposing a unified formulation in which each scalar value in the multivariate input window is treated as a node in a \emph{hypervariate graph}. Fourier Graph Operators (FGOs) then perform graph convolutions in the time domain by carrying out the equivalent matrix multiplications in Fourier space, achieving log-linear complexity and strong accuracy while eliminating the need for separate spatial and temporal modules.

\subsection{Time Series Forecasting with Spiking Neural Networks}

Within the SNN forecasting literature, Lv~et~al.~\cite{lv2024efficienteffectivetimeseriesforecasting} established the first systematic study by proposing spiking variants of TCN, RNN, and Transformer backbones, demonstrating competitive accuracy with significantly reduced energy consumption. However, sequence modeling in spiking Transformers faces a structural challenge: standard self-attention is permutation-invariant, and positional encoding under spike-based processing remains largely underexplored, which can weaken long-range dependency modeling. To mitigate this, subsequent work proposes spike-compatible positional encodings inspired by Central Pattern Generators (CPGs), injecting structured timing signals into spiking models~\cite{lv2024advancingspikingneuralnetworks}, though positional encoding in spiking architectures remains an open challenge.

A complementary direction improves the spiking neuron itself. The Temporal Segment LIF (TS-LIF) model~\cite{feng2025tsliftemporalsegmentspiking} introduces a dual-compartment neuron design in which dendritic and somatic pathways specialize in low- and high-frequency components respectively, improving multi-timescale integration and alleviating the long-horizon limitations of standard LIF dynamics. However, TS-LIF and the aforementioned spiking backbones process each variable independently and lack explicit mechanisms for capturing inter-variable correlations---a limitation the authors identify as future work.

Most closely related to our work, SpikF~\cite{wu2025spikf} demonstrates that Fourier-domain processing is a natural fit for SNN-based forecasting. By encoding patches of the input sequence and applying a Spiking Frequency Selection mechanism via a Spiking Fast Fourier Transform (S-FFT), SpikF avoids the permutation-invariance problem of self-attention while naturally exploiting the positional structure embedded in the Fourier transform. SpikF further provides theoretical efficiency analysis showing that S-FFT operations yield substantially lower energy consumption than their floating-point counterparts when spike trains are sparse~\cite{wu2025spikf,Lopez_Randulfe_2022,orchard2021efficientneuromorphicsignalprocessing}. Yet, like prior spiking methods, SpikF does not model cross-variable dependencies through an explicit graph structure.

Motivated by these advances, \textbf{SpikF-GO} bridges this gap by combining the unified hypervariate graph formulation of FourierGNN~\cite{yi2023fouriergnnrethinkingmultivariatetime} with spike-driven Fourier-domain processing inspired by SpikF~\cite{wu2025spikf}, bringing graph-based multivariate modeling into the spiking domain for multivariate TSF. We additionally present \textbf{SpikF-GO w/ CPG}, which injects CPG positional signals prior to spectral mixing~\cite{lv2024advancingspikingneuralnetworks} to strengthen long-range temporal modeling.

\section{Problem Formulation}

\subsection{Multivariate Time Series Forecasting}

We focus on the multivariate TSF setting, where the goal is to predict future values of multiple correlated variables over time.

A batch of multivariate time series with batch size $B$, sequence length $T$, and $N$ variables is represented as
\begin{equation}
\mathbf{X} \in \mathbb{R}^{B \times T \times N},
\end{equation}
where $\mathbf{X}_{b,t,n}$ denotes the value of variable $n\in\{1,\ldots,N\}$ at time $t\in\{1,\ldots,T\}$ for sample $b\in\{1,\ldots,B\}$, and we denote the $N$-dimensional observation vector by $\mathbf{x}_{b,t}\in\mathbb{R}^{N}$.

Given a look-back window length $L<T$ and a forecast horizon $O=T-L$, we construct, for each sample $b$, the input window and forecast horizon as
\begin{align}
\mathbf{X}^{(b)}_{\text{in}}
&=
\big[\mathbf{x}_{b,1}, \mathbf{x}_{b,2}, \ldots, \mathbf{x}_{b,L}\big]
\in \mathbb{R}^{L \times N},\\
\mathbf{Y}^{(b)}
&=
\big[\mathbf{x}_{b,L+1}, \mathbf{x}_{b,L+2}, \ldots, \mathbf{x}_{b,T}\big]
\in \mathbb{R}^{O \times N}.
\end{align}
Stacking $\{\mathbf{X}^{(b)}_{\text{in}}\}_{b=1}^{B}$ and $\{\mathbf{Y}^{(b)}\}_{b=1}^{B}$ yields the batched tensors
\begin{equation}
\mathbf{X}_{\text{in}} \in \mathbb{R}^{B \times L \times N},
\qquad
\mathbf{Y} \in \mathbb{R}^{B \times O \times N}.
\end{equation}

A forecasting model $\mathbf{F}_{\boldsymbol{\theta}}$ parameterized by $\boldsymbol{\theta}$ learns the mapping
\begin{equation}
\mathbf{F}_{\boldsymbol{\theta}}:\ \mathbb{R}^{B \times L \times N} \rightarrow \mathbb{R}^{B \times O \times N},
\qquad
\mathbf{X}_{\text{in}} \mapsto \mathbf{F}_{\boldsymbol{\theta}}(\mathbf{X}_{\text{in}}).
\end{equation}

The model is trained by minimizing the Mean Squared Error (MSE) over mini-batches of training samples, with model parameters updated iteratively via backpropagation through time (BPTT):
\begin{equation}
\label{eq:mse}
\mathcal{L}_{\text{MSE}}
=
\frac{1}{B\,O\,N}
\left\|
\mathbf{Y} - \mathbf{F}_{\boldsymbol{\theta}}(\mathbf{X}_{\text{in}})
\right\|_{F}^{2},
\end{equation}
where $\|\cdot\|_F$ denotes the Frobenius norm.

\subsection{Spiking Neuron Dynamics and Training}
The fundamental unit of our model is the Leaky Integrate-and-Fire (LIF) neuron~\cite{MAASS19971659,lv2024efficienteffectivetimeseriesforecasting}, whose membrane potential $U[t]$ evolves at each discrete time step $t$ as
\begin{align}
U[t] &= H[t-\Delta t] + I[t], \label{eq:lif_charge}\\
S[t] &= \mathbf{1}\bigl(U[t] \geq \vartheta\bigr), \label{eq:lif_spike}\\
H[t] &= V_{\text{reset}}\,S[t] + \bigl(1 - S[t]\bigr)\,\beta\, U[t], \label{eq:lif_reset}
\end{align}
where $\Delta t$ is the discretization constant controlling the granularity of LIF modeling, $I[t]$ is the input current computed by the preceding layer, $\beta < 1$ is the membrane decay factor, $\vartheta$ is the firing threshold, and $V_{\text{reset}}$ is the reset potential. When the membrane potential reaches $\vartheta$, the neuron emits a binary spike $S[t]=1$ and the potential is reset; otherwise it decays by $\beta$.

Since the Heaviside indicator $\mathbf{1}(\cdot)$ is non-differentiable, we adopt the arctangent surrogate gradient~\cite{doi:10.1126/sciadv.adi1480} during BPTT:
\begin{equation}
S[t] \approx \frac{1}{\pi}\arctan\!\left(\frac{\pi}{2}\,\alpha\, U[t]\right) + \frac{1}{2},
\end{equation}
where $\alpha$ controls the sharpness of the approximation.

\paragraph{Temporal Alignment.}
Following~\cite{lv2024efficienteffectivetimeseriesforecasting}, we align the continuous time series with the discrete spiking dimension by dividing each time-series step $\Delta T$ into $T_s$ finer SNN steps of size $\Delta t$, so that $\Delta T = T_s \Delta t$. This bridges the time-series time step $\Delta T$ and the SNN time step $\Delta t$, allowing both to share the same temporal meaning. The model therefore processes $T_s \times T \times N$ possible spike events per sample, and a spike encoder converts the floating-point inputs into spike trains of temporal resolution $T_s$.

\section{Methodology}

\subsection{Overview}

\begin{figure}[t]
    \centering
    \includegraphics[width=\textwidth]{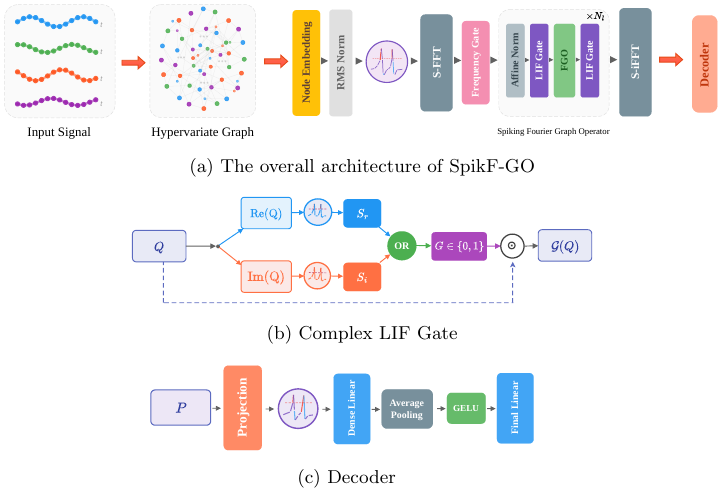}
    \caption{The overall architecture of SpikF-GO. (a) The input signal 
    is formed into a hypervariate graph, embedded and encoded into spike 
    trains, then processed by the S-FGO block with sparse frequency gating 
    and $N_{\ell}$ sequential Complex LIF-gated Fourier Graph Operators before 
    decoding. (b) The Complex LIF gate applies independent LIF neurons to 
    real and imaginary parts, combined via logical OR to produce a binary 
    spike mask. (c) The decoder compresses the temporal dimension, applies 
    a LIF layer with average pooling over SNN steps, followed by GELU 
    activation and a final linear projection.}
\label{fig:architecture}
\end{figure}

Given a multivariate input $\mathbf{X}_{\text{in}} \in \mathbb{R}^{B \times L \times N}$, the model proceeds through three stages: (1) an \textbf{Encoder} that constructs a hypervariate graph representation, embeds it into a latent space, and converts it into spike trains, (2) a \textbf{Spiking Fourier Graph Operator (S-FGO)} that performs sparse spectral mixing via a Hard Concrete frequency gate and LIF-gated complex linear operators, and (3) a \textbf{Decoder} that maps the processed representations to the prediction horizon $\hat{\mathbf{Y}} \in \mathbb{R}^{B \times O \times N}$. The overall architecture of SpikF-GO is illustrated in Figure~\ref{fig:architecture}.

\subsection{Encoder with Hypervariate Graph}

Following FourierGNN~\cite{yi2023fouriergnnrethinkingmultivariatetime}, we construct a hypervariate graph by treating every scalar observation in the input window as a node. Unlike conventional spatio-temporal approaches that model temporal and cross-variable dependencies through separate modules, the hypervariate graph connects any two variables at any two time steps, simultaneously encoding intra-series temporal dependencies, inter-series spatial dependencies, and time-varying cross-variable interactions within a single unified structure.

We define the graph formation operator $\mathcal{H}: \mathbb{R}^{B \times L \times N} \to \mathbb{R}^{B \times M}$, which flattens the spatial and temporal dimensions of each sample into $M = N \times L$ graph nodes:
\begin{equation}
    \mathbf{X}^{G} = \mathcal{H}(\mathbf{X}_{\text{in}}) \in \mathbb{R}^{B \times M},
\end{equation}
where $\mathcal{H}$ is applied independently to each sample in the batch and each entry $x^{G}_{b,m}$ corresponds to a single node in the fully-connected hypervariate graph. Each node is then projected into an $E$-dimensional embedding space via a learnable vector $\mathbf{e} \in \mathbb{R}^{1 \times E}$:
\begin{equation}
    \mathbf{V} = \mathbf{X}^{G}_{(\cdot)} \cdot \mathbf{e} \in \mathbb{R}^{B \times M \times E},
\end{equation}
where the multiplication broadcasts over the embedding dimension so that each node scales the shared embedding~$\mathbf{e}$. The node embeddings are refined by a learnable affine transform $A_{\mathrm{enc}}(\cdot)$ followed by Root Mean Square Normalization (RMSNorm)~\cite{zhang2019rootmeansquarelayer} over the node axis $M$:
\begin{equation}
\hat{\mathbf{V}} = \operatorname{RMSNorm}\!\left(A_{\mathrm{enc}}(\mathbf{V})\right) \in \mathbb{R}^{B \times M \times E}.
\end{equation}
where RMSNorm normalizes each channel independently over the node axis without mean subtraction, chosen over LayerNorm for its compatibility with neuromorphic hardware~\cite{abreu2025neuromorphicprinciplesefficientlarge}.

To produce a multi-step spike representation compatible with the SNN temporal dimension, $\hat{\mathbf{V}}$ is replicated across $T_s$ SNN steps and modulated by learnable per-step parameters $\gamma_t, \beta_t \in \mathbb{R}$:
\begin{equation}
    \mathbf{U}_t = \hat{\mathbf{V}} \cdot \gamma_t + \beta_t, \quad t = 1, \ldots, T_s.
\end{equation}
Each modulated signal is passed through a LIF encoder layer to produce binary spike trains, yielding the spike tensor:
\begin{equation}
    \mathbf{S} = \bigl[\operatorname{LIF}(\mathbf{U}_1),\;\ldots,\;\operatorname{LIF}(\mathbf{U}_{T_s})\bigr] \in \{0,1\}^{T_s \times B \times M \times E}.
\end{equation}

\subsection{Spiking Fourier Graph Operators (S-FGO)}

\paragraph{S-FFT.}
Motivated by the efficiency of Fourier-domain graph convolutions~\cite{yi2023fouriergnnrethinkingmultivariatetime} and the spiking FFT~\cite{wu2025spikf}, we apply the Spiking Fast Fourier Transform (S-FFT) along the node axis $M$ of the spike tensor. Since multiplication in the Fourier domain of the hypervariate graph is equivalent to graph convolution in the node domain~\cite{yi2023fouriergnnrethinkingmultivariatetime}, all subsequent spectral operations perform implicit graph mixing. The spike tensor is transformed as:
\begin{equation}
    \mathbf{Z} = \mathcal{SF}(\mathbf{S}) \in \mathbb{C}^{T_s \times B \times F \times E},
\end{equation}
where $\mathcal{SF}$ denotes the S-FFT applied independently at each SNN step and $F = \lfloor M/2 \rfloor + 1$ is the number of frequency bins.

\paragraph{Hard Concrete Frequency Gate.}
To encourage sparse frequency utilization and enable hardware-friendly inference with a fixed set of active bins, we apply a Hard Concrete gate~\cite{louizos2018learningsparseneuralnetworks} over the $F$ frequency bins. A learnable log-odds parameter $\log \alpha_f$ is maintained for each bin $f$. During training, the binary concrete distribution~\cite{maddison2017concretedistributioncontinuousrelaxation} is stretched to the $(\gamma, \zeta)$ interval with $\gamma < 0$ and $\zeta > 1$, and rectified via a hard-sigmoid:
\begin{equation}
    \bar{S}_f = \sigma\!\left(\frac{\log u - \log(1-u) + \log \alpha_f}{\tau}\right)\!(\zeta - \gamma) + \gamma, \quad u \sim \mathrm{Uniform}(0,1),
\end{equation}
\begin{equation}
        M_f = \min\!\left(1,\; \max\!\left(0,\; \bar{S}_f\right)\right),
\end{equation}
where $\tau$ is a temperature parameter. This stretching ensures that the gate can take exact zero and one values, folding the probability mass of the underlying continuous distribution onto those endpoints~\cite{louizos2018learningsparseneuralnetworks}. At inference, the stochasticity is removed and the gate is binarized as $M_f \leftarrow \mathbf{1}[\sigma(\log \alpha_f)(\zeta - \gamma) + \gamma > 0.5]$ to obtain a fixed binary frequency mask suitable for neuromorphic deployment. The gated spectrum is:
\begin{equation}
    \tilde{\mathbf{Z}} = \mathbf{Z} \odot \mathbf{M}, \quad \mathbf{M} \in [0,1]^{F},
\end{equation}
where $\odot$ broadcasts over $T_s$, $B$, and $E$. An $\ell_0$ sparsity penalty is added to the training objective to promote frequency pruning:
\begin{equation}\label{eq:spar}
  \mathcal{L}_{\ell_0} = \frac{1}{F}\sum_{f=1}^{F} \sigma\!\left(\log \alpha_f\right).
\end{equation}

\paragraph{S-FGO Block.}
The S-FGO block applies a sequence of $N_{\ell}$ complex-valued linear operators in the frequency domain, with each operator gated by a Complex LIF activation. Since each complex linear operator acts in the Fourier domain of the hypervariate graph, it performs learnable spectral graph mixing, corresponding to an implicit graph-convolution-like operation in the node domain \cite{yi2023fouriergnnrethinkingmultivariatetime}. Given the gated spectrum $\tilde{\mathbf{Z}} \in \mathbb{C}^{T_s \times B \times F \times E}$, each layer applies a complex affine normalization $\mathbf{A}^{(n)}$, followed by a Complex LIF gate $\mathcal{G}$, and a complex linear operator $\mathbf{W}^{(n)}$:
\begin{equation}
    \mathbf{Z}^{(n)} = \mathcal{G}\!\left(\mathbf{W}^{(n)} \cdot \mathcal{G}\!\left(\mathbf{A}^{(n)}(\mathbf{Z}^{(n-1)})\right)\right), \quad n = 1, \ldots, N_{\ell},
\end{equation}
where $\mathbf{Z}^{(0)} = \tilde{\mathbf{Z}}$. Residual connections with learnable scaling parameters are employed between layers to stabilize training. The Complex LIF gate $\mathcal{G}$ operates on a complex tensor $\mathbf{Q} = \mathbf{Q}_r + i\,\mathbf{Q}_i$ by applying independent LIF neurons to the real and imaginary parts and combining them via a logical OR:
\begin{equation}
    \mathcal{G}(\mathbf{Q}) = \mathbf{Q} \odot \Bigl[\mathbf{1}\!\left(\operatorname{LIF}(\mathbf{Q}_r) > 0\right) \;\vee\; \mathbf{1}\!\left(\operatorname{LIF}(\mathbf{Q}_i) > 0\right)\Bigr],
\end{equation}
ensuring that frequency components are gated in a binary, event-driven fashion throughout the block, preserving the sparse computational nature of the spiking framework.

\paragraph{S-iFFT.}
After the S-FGO block, the processed spectrum $\mathbf{Z}^{(N_{\ell})} \in \mathbb{C}^{T_s \times B \times F \times E}$ is mapped back to the node domain via the Spiking Inverse Fast Fourier Transform (S-iFFT):
\begin{equation}
    \mathbf{P} = \mathcal{SF}^{-1}\!\left(\mathbf{Z}^{(N_{\ell})}\right) \in \mathbb{R}^{T_s \times B \times N \times E \times L},
\end{equation}
where the $M$-dimensional node axis is separated back into the original variable and temporal dimensions.

\subsection{Decoder}

The decoder maps $\mathbf{P} \in \mathbb{R}^{T_s \times B \times N \times E \times L}$ to the final predictions $\hat{\mathbf{Y}} \in \mathbb{R}^{B \times O \times N}$. First, the temporal dimension $L$ is compressed to a small projection dimension $p \ll L$ via a linear projection $\mathbf{W}_p \in \mathbb{R}^{L \times p}$:
\begin{equation}
    \mathbf{P}'_t = \mathbf{P}_t \mathbf{W}_p \in \mathbb{R}^{B \times N \times E \times p},
    \quad t = 1, \ldots, T_s.
\end{equation}
After reshaping $\mathbf{P}'_t$ by merging the last two dimensions into $D = E \cdot p$, a LIF layer and a linear projection $\mathbf{W}_1 \in \mathbb{R}^{D \times d_r}$ are applied at each spiking step. The resulting representations are averaged over the $T_s$ SNN steps, followed by a Gaussian Error Linear Unit (GELU) activation and a final linear projection $\mathbf{W}_2 \in \mathbb{R}^{d_r \times O}$. Since the linear layers act on the last dimension, the decoder first produces outputs in $\mathbb{R}^{B \times N \times O}$, which are then transposed to match the forecasting convention:
\begin{equation}
    \hat{\mathbf{Y}} =
    \mathrm{Transpose}_{(N,O)}\!\left(
    \mathrm{GELU}\!\left(
    \frac{1}{T_s}\sum_{t=1}^{T_s}
    \mathrm{LIF}\!\left(\mathrm{Reshape}(\mathbf{P}'_t)\right)\mathbf{W}_1
    \right)\mathbf{W}_2
    \right).
\end{equation}
Here, $\mathbf{W}_1$ and $\mathbf{W}_2$ are weight-normalized linear projections, and $d_r$ is a reduced hidden dimension.

\section{Experiments}

\subsection{Experimental Settings}

\paragraph{Datasets.}
We evaluate on eight public multivariate TSF benchmarks spanning traffic flow (Traffic, METR-LA, PEMS-BAY), energy systems (Solar, Electricity), epidemiological records (COVID-19), biomedical signals (ECG), and web activity (Wiki), with $55$--$2{,}000$ variables and 5-minute to daily granularity, providing a broad evaluation setting for TSF \cite{sen2019thinkgloballyactlocally,yi2023fouriergnnrethinkingmultivariatetime}; summary statistics are in Appendix~\ref{app:A}. Following~\cite{yi2023fouriergnnrethinkingmultivariatetime}, all datasets except COVID-19 are split chronologically into training, validation, and test sets with a 7:2:1 ratio; for COVID-19 the ratio is 6:2:2.

\paragraph{Baselines.}
We evaluate the proposed \textbf{SpikF-GO} model introduced in Section~4 and a
variant, \textbf{SpikF-GO w/ CPG}, which uses the same CPG-based positional
encoding as in~\cite{lv2024advancingspikingneuralnetworks}. The CPG module is
used exactly as implemented in the original work. We compare these models against
nine SNN baselines spanning the major architectural families in SNN-based
forecasting. These include SpikeRNN and SpikeTCN with CPG positional encodings,
as well as Spikformer w/ CPG~\cite{lv2024advancingspikingneuralnetworks};
TS-GRU, TS-TCN, and TS-Former~\cite{feng2025tsliftemporalsegmentspiking};
SpikF~\cite{wu2025spikf}; and Spike-GRU and
iSpikformer~\cite{lv2024efficienteffectivetimeseriesforecasting}. This set includes strong recent SNN models such as the TS-LIF-based models and SpikF,
whose original study reports competitive or better average performance than other ANN
baselines such as iTransformer \cite{liu2024itransformerinvertedtransformerseffective}. The single ANN
baseline is FourierGNN~\cite{yi2023fouriergnnrethinkingmultivariatetime}, which
is among the state-of-the-art graph-based ANN forecasters for multivariate TSF
and the closest continuous-valued counterpart to SpikF-GO, since both methods
build on the hypervariate graph formulation. For each included baseline, we use the hyperparameters reported
in the original paper on the datasets originally evaluated; for additional
datasets, hyperparameters are tuned on the validation set. Unless stated
otherwise, our model uses $N_{\ell}=3$ and $E=128$, matching FourierGNN.

\paragraph{Training Settings.}
All experiments use PyTorch~2.5.1 on a single NVIDIA RTX~4090. All models are trained with MSE as the main objective (Eq.~\ref{eq:mse}). For SpikF-GO, we additionally apply an adaptive $\ell_0$ sparsity penalty to the frequency gate (Eq.~\ref{eq:spar}) to encourage sparse frequency utilization. We apply Reversible Instance Normalization (RevIN)~\cite{kim2021reversible} to all models and correct a data leakage issue in the original FourierGNN codebase, where normalization statistics were computed over the full dataset rather than using training data only. In our experiments, preprocessing is performed strictly using training-set statistics, ensuring proper train–test separation. All SNN models use $T_s=4$ spiking time steps, except SpikF, which uses $T_s=16$, following the original setting. We evaluate test set performance using the Coefficient of Determination ($R^2$) and Mean Absolute Error (MAE); their formal definitions are provided in Appendix \ref{app:B}.

\subsection{Main Results}

Tables~\ref{tab:results1} and~\ref{tab:results2} report forecasting results on eight benchmarks, averaged over 5 runs, using $R^2$ (higher is better) and MAE (lower is better). The input window and forecast horizon are both 12. \textbf{Bold} and \underline{underline} denote the best and second-best results, respectively. Avg.\ Rank is the mean rank over all eight datasets, computed separately for each metric (lower is better).

SpikF-GO w/ CPG achieves the best overall average rank, ranking first on both \textbf{$R^2$ (2.4)} and MAE \textbf{(2.3)}, while SpikF-GO achieves the second-best average rank on \textbf{$R^2$ (2.8)}. The benefit of CPG is consistent with findings in \cite{lv2024advancingspikingneuralnetworks}, where injecting explicit positional structure improved spiking RNN, TCN, and Transformer backbones; here, CPG similarly strengthens SpikF-GO on most datasets. Among the baselines, SpikF is the strongest SNN baseline, while FourierGNN achieves the second-best average rank on MAE. The remaining SNN baselines perform notably worse, likely due to the lack of explicit cross-variate modeling. Results with standard deviations and per-horizon breakdowns are provided in Appendix \ref{app:C}.

\begin{table}[!t]
\centering
\caption{Forecasting results on ECG, COVID-19 (COVID), Solar, and Electricity (ECL), averaged over 5 runs. \textbf{Bold} and \underline{underline} indicate the best and second-best results in each column, respectively. Avg.\ Rank is computed over all eight datasets.}
\label{tab:results1}
\arrayrulecolor{black}

\begin{tabular}{c:c:cccc:c}
\hline
\textbf{Models} & \textbf{Metric} & \textbf{ECG} & \textbf{COVID} & \textbf{Solar} & \textbf{ECL} & \textbf{Avg. Rank}$\downarrow$ \\
\hline

\multirow{2}{*}{FourierGNN }
  & $R^{2}\!\!\uparrow$  & \underline{.707} & .950 & \textbf{.742} & .912 & 4.8 \\
  & MAE$\downarrow$ & .329 & 31.50 & \textbf{9.03} & 97.13 & \underline{3.1} \\
\hline
\hline

\multirow{2}{*}{Spike-GRU }
  & $R^{2}\!\!\uparrow$  & .706 & .845 & .487 & .898 & 10.5 \\
  & MAE$\downarrow$ & .331 & 55.35 & 14.34 & 107.09 & 10.7 \\

\multirow{2}{*}{iSpikformer }
  & $R^{2}\!\!\uparrow$  & \underline{.707} & .874 & .593 & .829 & 7.7 \\
  & MAE$\downarrow$ & .330 & 49.79 & 12.27 & 139.73 & 8.1 \\
\hline

\multirow{2}{*}{SpikeRNN w/ CPG }
  & $R^{2}\!\!\uparrow$  & \underline{.707} & .836 & .680 & .916 & 8.3 \\
  & MAE$\downarrow$ & .330 & 57.87 & 10.37 & 96.45 & 8.8 \\

\multirow{2}{*}{SpikeTCN w/ CPG }
  & $R^{2}\!\!\uparrow$  & \underline{.707} & .849 & .706 & .910 & 6.4 \\
  & MAE$\downarrow$ & .330 & 55.20 & 9.84 & 96.17 & 6.9 \\

\multirow{2}{*}{Spikformer w/ CPG }
  & $R^{2}\!\!\uparrow$  & .706 & .867 & .699 & \underline{.930} & 6.5 \\
  & MAE$\downarrow$ & .330 & 48.71 & 10.03 & \textbf{84.65} & 6.2 \\
\hline

\multirow{2}{*}{TS-GRU }
  & $R^{2}\!\!\uparrow$  & \underline{.707} & .842 & .590 & .902 & 8.9 \\
  & MAE$\downarrow$ & .329 & 56.81 & 12.13 & 103.36 & 7.6 \\

\multirow{2}{*}{TS-TCN }
  & $R^{2}\!\!\uparrow$  & \underline{.707} & .836 & .707 & .893 & 7.4 \\
  & MAE$\downarrow$ & .330 & 57.89 & 9.80 & 104.51 & 8.8 \\

\multirow{2}{*}{TS-Former }
  & $R^{2}\!\!\uparrow$  & .704 & .835 & .679 & .925 & 7.9 \\
  & MAE$\downarrow$ & .330 & 58.02 & 10.50 & \underline{84.78} & 7.6 \\
\hline

\multirow{2}{*}{SpikF}
  & $R^{2}\!\!\uparrow$  & \textbf{.708} & .961 & .712 & .910 & 4.4 \\
  & MAE$\downarrow$ & \underline{.327} & 28.16 & 9.61 & 99.99 & 4.3 \\
\hline
\hline

\rowcolor{blue!9}
\textbf{SpikF-GO}
  & $R^{2}\!\!\uparrow$  & \textbf{.708} & \textbf{.967} & \underline{.740} & .927 & \underline{2.8} \\
\rowcolor{blue!9}
  & MAE$\downarrow$ & \underline{.327} & \textbf{25.15} & \underline{9.18} & 91.15 & 3.8 \\

\rowcolor{blue!9}
\textbf{SpikF-GO w/ CPG}
  & $R^{2}\!\!\uparrow$  & \textbf{.708} & \underline{.965} & \textbf{.742} & \textbf{.942} & \textbf{2.4} \\
\rowcolor{blue!9}
  & MAE$\downarrow$ & \textbf{.326} & \underline{26.09} & \textbf{9.03} & 89.14 & \textbf{2.3} \\
\hline
\end{tabular}
\end{table}

\begin{table}[!t]
\centering
\caption{Forecasting results on METR-LA (METR), Traffic, PEMS-BAY (PEMS), and Wiki, averaged over 5 runs. \textbf{Bold} and \underline{underline} denote the best and second-best results in each column. Avg.\ Rank is computed over all eight datasets.}
\label{tab:results2}
\arrayrulecolor{black}

\begin{tabular}{c:c:cccc:c}
\hline
\textbf{Models} & \textbf{Metric} & \textbf{METR} & \textbf{Traffic} & \textbf{PEMS} & \textbf{Wiki} & \textbf{Avg. Rank}$\downarrow$ \\
\hline

\multirow{2}{*}{FourierGNN }
  & $R^{2}\!\!\uparrow$  & \underline{.766} & .632 & .733 & .473 & 4.8 \\
  & MAE$\downarrow$ & \textbf{.071} & .013 & \textbf{2.16} & \textbf{185.64} & \underline{3.1}\\
\hline
\hline

\multirow{2}{*}{Spike-GRU }
  & $R^{2}\!\!\uparrow$  & .691 & .491 & .573 & .458 & 10.5 \\
  & MAE$\downarrow$ & .086 & .020 & 2.91 & 201.08 & 10.7 \\

\multirow{2}{*}{iSpikformer }
  & $R^{2}\!\!\uparrow$  & .748 & .483 & .739 & .469 & 7.7 \\
  & MAE$\downarrow$ & .075 & .018 & 2.25 & 195.26 & 8.1 \\
\hline

\multirow{2}{*}{SpikeRNN w/ CPG }
  & $R^{2}\!\!\uparrow$  & .728 & .553 & .732 & .458 & 8.3 \\
  & MAE$\downarrow$ & .079 & .017 & 2.30 & 202.26 & 8.8 \\

\multirow{2}{*}{SpikeTCN w/ CPG }
  & $R^{2}\!\!\uparrow$  & .759 & .688 & .747 & .447 & 6.4 \\
  & MAE$\downarrow$ & .073 & .014 & 2.25 & 218.76 & 6.9 \\

\multirow{2}{*}{Spikformer w/ CPG}
  & $R^{2}\!\!\uparrow$  & .747 & \underline{.728} & .721 & .467 & 6.5 \\
  & MAE$\downarrow$ & .076 & \underline{.012} & 2.28 & 202.63 & 6.2 \\
\hline

\multirow{2}{*}{TS-GRU }
  & $R^{2}\!\!\uparrow$  & .758 & .437 & .735 & .451 & 8.9 \\
  & MAE$\downarrow$ & .074 & .019 & 2.25 & 197.42 & 7.6 \\

\multirow{2}{*}{TS-TCN }
  & $R^{2}\!\!\uparrow$  & .692 & .689 & .743 & .459 & 7.4 \\
  & MAE$\downarrow$ & .086 & .014 & 2.27 & 204.46 & 8.8 \\

\multirow{2}{*}{TS-Former }
  & $R^{2}\!\!\uparrow$  & .747 & \textbf{.744} & .723 & .465 & 7.9 \\
  & MAE$\downarrow$ & .076 & \textbf{.011} & 2.29 & 206.41 & 7.6 \\
\hline

\multirow{2}{*}{SpikF}
  & $R^{2}\!\!\uparrow$  & \underline{.766} & .541 & .740 & \underline{.477} & 4.4 \\
  & MAE$\downarrow$ & \underline{.072} & .016 & 2.21 & \underline{186.58} & 4.3 \\
\hline
\hline

\rowcolor{blue!9}
\textbf{SpikF-GO}
  & $R^{2}\!\!\uparrow$ & .762 & .651 & \underline{.762} & \textbf{.480} & \underline{2.8} \\
\rowcolor{blue!9}
  & MAE$\downarrow$ & .077 & .013 & \underline{2.17} & 189.22 & 3.8 \\

\rowcolor{blue!9}
\textbf{SpikF-GO w/ CPG}
  & $R^{2}\!\!\uparrow$ & \textbf{.769} & .669 & \textbf{.766} & .466 & \textbf{2.4} \\
\rowcolor{blue!9}
  & MAE$\downarrow$ & \underline{.072} & \underline{.012} & \textbf{2.16} & 191.21 & \textbf{2.3} \\
\hline
\end{tabular}
\end{table}

\subsection{Model Analysis}

We ablate three components on Solar, Traffic, and COVID-19, as shown in Table~\ref{tab:ablation}. The \textbf{Temporal-Only} variant, which removes cross-variable modeling by processing each variable independently, causes the largest drop across all datasets, with $R^2$ decreasing by 0.018, 0.074, and 0.030, respectively. This confirms that the hypervariate graph formulation is the main source of SpikF-GO's performance gains, particularly on Traffic, where sensor correlations are strong. Replacing the learned Hard Concrete gate with fixed \textbf{Top-K} frequency selection also degrades performance, indicating that adaptive frequency pruning is more effective than a fixed strategy. Replacing RMSNorm with a simple \textbf{Scale-Shift} affine transform, which is more suitable for neuromorphic hardware, yields nearly identical performance on Solar and only slight degradation on Traffic and COVID-19, making it a practical hardware-friendly alternative.

\begin{table}[!t]
\centering
\caption{Ablation study on key components of SpikF-GO. Best results are shown in \textbf{bold}.}
\label{tab:ablation}
\begin{tabular}{lcccccc}
\toprule
 & \multicolumn{2}{c}{\textbf{Solar}} & \multicolumn{2}{c}{\textbf{Traffic}} & \multicolumn{2}{c}{\textbf{COVID-19}} \\
\cmidrule(lr){2-3} \cmidrule(lr){4-5} \cmidrule(lr){6-7}
Variant & $R^2 \uparrow$ & MAE $\downarrow$ & $R^2 \uparrow$ & MAE $\downarrow$ & $R^2 \uparrow$ & MAE $\downarrow$ \\
\midrule
SpikF-GO              & \textbf{0.743} & \textbf{9.15} & \textbf{0.621} & \textbf{0.014} & \textbf{0.967} & \textbf{24.87} \\
Temporal-Only         & 0.725 & 9.42 & 0.547 & 0.015 & 0.937 & 34.50 \\
Top-K Gate            & 0.728 & 9.66 & 0.619   & \textbf{0.014}   & 0.955 & 29.15 \\
Scale-Shift           &\textbf{0.743}   & \textbf{9.15} & 0.620    & \textbf{0.014} & 0.964 & 26.06\\
\bottomrule
\end{tabular}
\end{table}

\begin{figure}[!t]
\centering
\includegraphics[width=\textwidth]{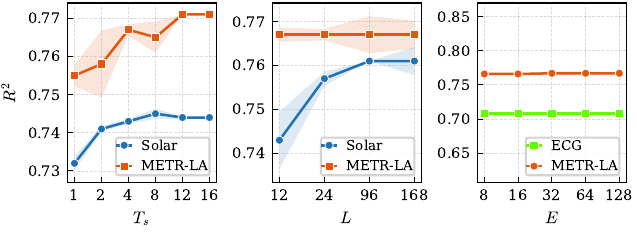}
\caption{Sensitivity analysis of SpikF-GO on three hyperparameters:
spiking timesteps~$T_s$, input window length~$L$, and
embedding size~$E$. Shaded regions denote $\pm$1 std over three runs.}
\label{fig:ablation}
\end{figure}

Figure~\ref{fig:ablation} shows the sensitivity of SpikF-GO to three hyperparameters: spiking timesteps~$T_s$, input window length~$L$, and embedding size~$E$. Shaded regions denote $\pm$1 standard deviation over three runs. Performance improves with increasing~$T_s$, peaking at $T_s{=}8$ for Solar and $T_s{=}12$ for METR-LA, with no further gains thereafter. Increasing the input window from $L{=}96$ to $L{=}168$ yields negligible improvement, indicating that $L{=}96$ is sufficient for the considered datasets. Likewise, $R^2$ remains nearly unchanged from $E{=}8$ to $E{=}128$, showing that SpikF-GO maintains performance even with compact embeddings, which is desirable for neuromorphic deployment due to lower energy and memory costs.

Table~\ref{tab:energy-full} reports theoretical energy consumption, wall-clock runtime, and energy reduction relative to FourierGNN on Solar with prediction length 12. Energy is estimated on 45\,nm hardware following~\cite{Lemaire_2023} and decomposed into memory-access ($E_{\mathrm{Mem}}$), operational ($E_{\mathrm{Ops}}$), and addressing ($E_{\mathrm{Addr}}$) components. The energy cost of S-FFT and S-iFFT is estimated following~\cite{wu2025spikf}. We assume 4.6\,pJ per FLOP and 0.9\,pJ per synaptic operation (SOP) on 45\,nm hardware~\cite{yao2022attentionspikingneuralnetworks}. Training and inference times are reported as average per-batch runtimes.

As shown in Table~\ref{tab:energy-full}, SpikF-GO reduces energy consumption by $1.89\times$ relative to FourierGNN while achieving better forecasting performance. Reducing the embedding dimension to $E{=}8$ further increases the reduction to $7.86\times$ with little change in performance (Fig.~\ref{fig:ablation}), making it the most energy-efficient graph-based configuration. Although SpikF achieves lower energy consumption than SpikF-GO with $E{=}128$ ($4.27\times$ relative to FourierGNN), it does not explicitly model cross-variate dependencies and does not outperform the full SpikF-GO model. In terms of wall-clock runtime, FourierGNN is the fastest because it relies only on ANN computation, whereas SNN-based models incur additional overhead from multi-step spiking simulation on GPU; this overhead would not arise on neuromorphic hardware.

\begin{table}[!t]
\centering
\caption{Energy consumption and runtime comparison across models. $\downarrow\!n\!\times$ denotes the energy reduction factor relative to the ANN baseline FourierGNN. Runtime is reported as training\,/\,inference time per batch in seconds.}
\label{tab:energy-full}

\begin{tabular}{lccccc}
\toprule
\textbf{Model}
  & $E_{\mathrm{Mem}}$/$\mu$J
  & $E_{\mathrm{Ops}}$/$\mu$J
  & $E_{\mathrm{Addr}}$/$\mu$J
  & $E_{\mathrm{Total}}$/$\mu$J
  & Time/s \\
\midrule
FourierGNN
  & $7.76 \times 10^{4}$
  & $1.13 \times 10^{4}$
  & $1.29 \times 10^{1}$
  & $8.89 \times 10^{4}$
  & 0.012\,/\,0.002 \\
\midrule
SpikF-GO
  & $4.43 \times 10^{4}$
  & $2.17 \times 10^{3}$
  & $5.19 \times 10^{2}$
  & $4.70 \times 10^{4}$ $\downarrow 1.89\times$
  & 0.027\,/\,0.011 \\
SpikF
  & $2.05 \times 10^{4}$
  & $2.61 \times 10^{2}$
  & $7.91 \times 10^{1}$
  & $2.08 \times 10^{4}$ $\downarrow 4.27\times$
  & 0.077\,/\,0.037 \\
SpikF-GO ($E\!=\!8$)
  & $9.28 \times 10^{3}$
  & $2.00 \times 10^{3}$
  & $2.79 \times 10^{1}$
  & $\mathbf{1.13 \times 10^{4}}$ $\downarrow \mathbf{7.86}\times$
  & 0.020\,/\,0.010 \\
\bottomrule
\end{tabular}
\end{table}

\section{Conclusion}

We introduced SpikF-GO, a spiking model for multivariate time series forecasting that addresses the lack of explicit cross-variable modeling in prior SNN forecasting methods through a hypervariate graph formulation and spike-driven Fourier-domain graph processing. We further presented SpikF-GO w/ CPG, which strengthens long-range temporal modeling through positional encoding. Experiments on eight benchmark datasets demonstrated the effectiveness of the proposed models, with SpikF-GO w/ CPG achieving the best overall average rank and SpikF-GO also outperforming strong SNN and ANN baselines. In addition, the proposed approach provides clear theoretical energy benefits over FourierGNN, highlighting its potential for efficient neuromorphic forecasting. 

At the same time, our method has several limitations. Its complexity and energy consumption increase when both the input length and the number of channels become large, which may limit scalability in high-dimensional long-horizon settings. Moreover, our energy analysis is theoretical and does not yet include deployment on real neuromorphic hardware. Future work will therefore focus on improving scalability, reducing computational cost for large-scale inputs, and validating the proposed models in practical neuromorphic deployment scenarios.

\begin{credits}
\subsubsection{\discintname}
The authors have no competing interests to declare.

\subsubsection{Use of Generative AI.}
Generative AI was used only for grammar correction and language polishing.
\end{credits}

\bibliographystyle{splncs04}
\bibliography{references}

\clearpage

\appendix
\renewcommand{\theHsection}{appendix.\Alph{section}}
\renewcommand{\theHsubsection}{appendix.\Alph{section}.\arabic{subsection}}
\phantomsection
\section{Dataset Statistics}\label{app:A}

Table~\ref{tab:datasets} summarizes the benchmark datasets used in our experiments, including the number of variables or channels (\#Vars.), the number of observations (\#Obs.), the sampling frequency (Freq.), and the application domain. For the Wiki dataset, we randomly sample 2,000 variables from the full dataset, which contains more than 100,000 time series. We provide the processed datasets used in our experiments at \url{https://figshare.com/s/7617530bce306584fe95?file=62576929}.

\begin{table}[ht]
\centering
\caption{Summary of benchmark datasets. ECL denotes Electricity.}
\label{tab:datasets}
\begin{tabular}{l c c c c c c c c}
\toprule
 & ECG & COVID-19 & Solar & ECL & Traffic & METR-LA & PEMS-BAY & Wiki \\
\midrule
\#Vars. & 140 & 55 & 593 & 370 & 963 & 207 & 325 & 2\,000 \\
\#Obs.  & 5\,000 & 335 & 3\,650 & 140\,211 & 10\,560 & 34\,272 & 52\,116 & 803 \\
Freq.   & -- & 1\,day & 1\,hr & 15\,min & 1\,hr & 5\,min & 5\,min & 1\,day \\
Domain  & Bio. & Epid. & Energy & Energy & Traffic & Traffic & Traffic & Web \\
\bottomrule
\end{tabular}
\end{table}

\phantomsection
\section{Evaluation Metrics}\label{app:B}

We evaluate all models using two complementary metrics: the Coefficient of Determination~($R^2$) and Mean Absolute Error~(MAE). Following the notation introduced in Section~3.1, let $N$ denote the number of variables, $O$ the forecasting horizon, and $\hat{\mathbf{Y}} = F_\theta(\mathbf{X}_{\text{in}}) \in \mathbb{R}^{B \times O \times N}$ the model predictions. We write $Y^{b}_{o,n}$ and $\hat{Y}^{b}_{o,n}$ for the ground-truth and predicted value of variable~$n$ at horizon step~$o$ for sample~$b$, respectively.

\paragraph{Coefficient of Determination ($R^2$).}
$R^2$ measures the proportion of variance in the ground-truth targets explained by the model. Higher values indicate better fit, with a maximum of~$1$.
\begin{equation}
R^2
=
1
-
\frac{\displaystyle\sum_{b=1}^{B}\sum_{o=1}^{O}\sum_{n=1}^{N}
\left(Y^{b}_{o,n}-\hat{Y}^{b}_{o,n}\right)^2}
{\displaystyle\sum_{b=1}^{B}\sum_{o=1}^{O}\sum_{n=1}^{N}
\left(Y^{b}_{o,n}-\bar{Y}\right)^2},
\end{equation}
where
\[
\bar{Y}=\frac{1}{B O N}\sum_{b=1}^{B}\sum_{o=1}^{O}\sum_{n=1}^{N} Y^{b}_{o,n}
\]
is the global mean of the ground-truth targets over the test set.

\paragraph{Mean Absolute Error (MAE).}
MAE measures the average absolute difference between the predicted and ground-truth values. Lower values indicate better performance.
\begin{equation}
\mathrm{MAE}
=
\frac{1}{B O N}
\sum_{b=1}^{B}\sum_{o=1}^{O}\sum_{n=1}^{N}
\left|Y^{b}_{o,n}-\hat{Y}^{b}_{o,n}\right|.
\end{equation}

\phantomsection
\section{Additional Results}\label{app:C}

This section provides additional experimental results. Section~\ref{app:C1} presents the main-paper results with standard deviations, and Section~\ref{app:C2} reports results across forecast horizons $\{6,12,24,48\}$, also with standard deviations.

\phantomsection
\subsection{Main Results with Standard Deviations}\label{app:C1}

Tables~\ref{tab:app-results1} and~\ref{tab:app-results2} report the results for the setting with input length 12 and forecast horizon 12. In addition to the main-paper results, they also include the corresponding mean and standard deviation over 5 random seeds.

\begin{table}[ht]
\centering
\caption{Forecasting results on ECG, COVID-19 (COVID), Solar, and Electricity (ECL), averaged over 5 runs. Values are reported as mean $\pm$ standard deviation. \textbf{Bold} and \underline{underline} indicate the best and second-best results in each column, respectively.}
\label{tab:app-results1}
\providecommand{\sd}[1]{{$\pm$#1}}
\begin{tabular}{c:c:cccc}
\hline
\textbf{Models} & \textbf{Metric} & \textbf{ECG} & \textbf{COVID} & \textbf{Solar} & \textbf{ECL} \\
\hline
\multirow{2}{*}{FourierGNN}
  & $R^{2}\!\!\uparrow$ & \underline{.707}\sd{.0002} & .950\sd{.0056} & \textbf{.742}\sd{.0010} & .912\sd{.0013} \\
  & MAE$\downarrow$     & .329\sd{.0001} & 31.50\sd{1.575} & \textbf{9.03}\sd{.0193} & 97.13\sd{.474} \\
\hline
\hline
\multirow{2}{*}{Spike-GRU}
  & $R^{2}\!\!\uparrow$ & .706\sd{.0002} & .845\sd{.0036} & .487\sd{.0045} & .898\sd{.0016} \\
  & MAE$\downarrow$     & .331\sd{.0002} & 55.35\sd{.839} & 14.34\sd{.1003} & 107.09\sd{1.04} \\
\addlinespace[1pt]
\multirow{2}{*}{iSpikformer }
  & $R^{2}\!\!\uparrow$ & \underline{.707}\sd{.0002} & .874\sd{.0237} & .593\sd{.0755} & .829\sd{.0195} \\
  & MAE$\downarrow$     & .330\sd{.0006} & 49.79\sd{5.166} & 12.27\sd{1.6421} & 139.73\sd{9.25} \\
\addlinespace[1pt]
\hline
\multirow{2}{*}{SpikeRNN w/ CPG}
  & $R^{2}\!\!\uparrow$ & \underline{.707}\sd{.0000} & .836\sd{.0031} & .680\sd{.0004} & .916\sd{.0037} \\
  & MAE$\downarrow$     & .330\sd{.0004} & 57.87\sd{.458} & 10.37\sd{.0862} & 96.45\sd{1.21} \\
\addlinespace[1pt]
\multirow{2}{*}{SpikeTCN w/ CPG}
  & $R^{2}\!\!\uparrow$ & \underline{.707}\sd{.0004} & .849\sd{.0068} & .706\sd{.0061} & .910\sd{.0059} \\
  & MAE$\downarrow$     & .330\sd{.0006} & 55.20\sd{1.964} & 9.84\sd{.0984} & 96.17\sd{3.26} \\
\addlinespace[1pt]
\multirow{2}{*}{Spikformer w/ CPG }
  & $R^{2}\!\!\uparrow$ & .706\sd{.0008} & .867\sd{.0022} & .699\sd{.0033} & \underline{.930}\sd{.0057} \\
  & MAE$\downarrow$     & .330\sd{.0006} & 48.71\sd{.489} & 10.03\sd{.1568} & \textbf{84.65}\sd{.429} \\
\addlinespace[1pt]
\hline
\multirow{2}{*}{TS-GRU }
  & $R^{2}\!\!\uparrow$ & \underline{.707}\sd{.0000} & .842\sd{.0061} & .590\sd{.0212} & .902\sd{.0021} \\
  & MAE$\downarrow$     & .329\sd{.0001} & 56.81\sd{.924} & 12.13\sd{.3755} & 103.36\sd{1.10} \\
\addlinespace[1pt]
\multirow{2}{*}{TS-TCN }
  & $R^{2}\!\!\uparrow$ & \underline{.707}\sd{.0000} & .836\sd{.0021} & .707\sd{.0071} & .893\sd{.0387} \\
  & MAE$\downarrow$     & .330\sd{.0003} & 57.89\sd{.436} & 9.80\sd{.1001} & 104.51\sd{21.02} \\
\addlinespace[1pt]
\multirow{2}{*}{TS-Former}
  & $R^{2}\!\!\uparrow$ & .704\sd{.0012} & .835\sd{.0002} & .679\sd{.0231} & .925\sd{.0086} \\
  & MAE$\downarrow$     & .330\sd{.0009} & 58.02\sd{.017} & 10.50\sd{.4997} & \underline{84.78}\sd{.615} \\
\addlinespace[1pt]
\hline
\multirow{2}{*}{SpikF }
  & $R^{2}\!\!\uparrow$ & \textbf{.708}\sd{.0001} & .961\sd{.0062} & .712\sd{.0012} & .910\sd{.0011} \\
  & MAE$\downarrow$     & \underline{.327}\sd{.0001} & 28.16\sd{1.802} & 9.61\sd{.0324} & 99.99\sd{.115} \\
\hline
\hline
\rowcolor{blue!9}
{\textbf{SpikF-GO}}
  & $R^{2}\!\!\uparrow$  & \textbf{.708}\sd{.0002} & \textbf{.967}\sd{.0042} & \underline{.740}\sd{.0048} & .927\sd{.0068} \\
\rowcolor{blue!9}
  & MAE$\downarrow$      & \underline{.327}\sd{.0001} & \textbf{25.15}\sd{1.654} & \underline{9.18}\sd{.0648} & 91.15\sd{1.29} \\
\addlinespace[1pt]

\rowcolor{blue!9}
{\textbf{SpikF-GO w/ CPG}}
  & $R^{2}\!\!\uparrow$  & \textbf{.708}\sd{.0005} & \underline{.965}\sd{.0024} & \textbf{.742}\sd{.0051} & \textbf{.942}\sd{.0028} \\
\rowcolor{blue!9}
  & MAE$\downarrow$      & \textbf{.326}\sd{.0003} & \underline{26.09}\sd{.703} & \textbf{9.03}\sd{.1060} & 89.14\sd{1.08} \\
\hline
\end{tabular}
\end{table}

\begin{table}[ht]
\centering
\caption{Forecasting results on METR-LA (METR), Traffic, PEMS-BAY (PEMS), and Wiki, averaged over 5 runs. Values are reported as mean $\pm$ standard deviation. \textbf{Bold} and \underline{underline} indicate the best and second-best results in each column, respectively.}
\label{tab:app-results2}
\begin{tabular}{c:c:cccc}
\hline
\textbf{Models} & \textbf{Metric} & \textbf{METR} & \textbf{Traffic} & \textbf{PEMS} & \textbf{Wiki} \\
\midrule
\multirow{2}{*}{FourierGNN}
  & $R^{2}\!\!\uparrow$ & \underline{.766}\sd{.0003} & .632\sd{.0032} & .733\sd{.0032} & .473\sd{.0063} \\
  & MAE$\downarrow$     & \textbf{.071}\sd{.0001} & .013\sd{.0001} & \textbf{2.16}\sd{.0023} & \textbf{185.64}\sd{.81} \\
\hline
\hline
\multirow{2}{*}{Spike-GRU}
  & $R^{2}\!\!\uparrow$ & .691\sd{.0022} & .491\sd{.0452} & .573\sd{.0033} & .458\sd{.0017} \\
  & MAE$\downarrow$     & .086\sd{.0004} & .020\sd{.0004} & 2.91\sd{.0201} & 201.08\sd{1.18} \\
\addlinespace[1pt]
\multirow{2}{*}{iSpikformer}
  & $R^{2}\!\!\uparrow$ & .748\sd{.0175} & .483\sd{.0124} & .739\sd{.0059} & .469\sd{.0058} \\
  & MAE$\downarrow$     & .075\sd{.0028} & .018\sd{.0004} & 2.25\sd{.0298} & 195.26\sd{1.92} \\
\addlinespace[1pt]
\hline
\multirow{2}{*}{SpikeRNN w/ CPG}
  & $R^{2}\!\!\uparrow$ & .728\sd{.0088} & .553\sd{.0376} & .732\sd{.0020} & .458\sd{.0010} \\
  & MAE$\downarrow$     & .079\sd{.0016} & .017\sd{.0009} & 2.30\sd{.0079} & 202.26\sd{1.26} \\
\addlinespace[1pt]
\multirow{2}{*}{SpikeTCN w/ CPG }
  & $R^{2}\!\!\uparrow$ & .759\sd{.0012} & .688\sd{.0166} & .747\sd{.0008} & .447\sd{.0076} \\
  & MAE$\downarrow$     & .073\sd{.0002} & .014\sd{.0008} & 2.25\sd{.0017} & 218.76\sd{5.52} \\
\addlinespace[1pt]
\multirow{2}{*}{Spikformer w/ CPG}
  & $R^{2}\!\!\uparrow$ & .747\sd{.0030} & \underline{.728}\sd{.0109} & .721\sd{.0046} & .467\sd{.0032} \\
  & MAE$\downarrow$     & .076\sd{.0006} & \underline{.012}\sd{.0002} & 2.28\sd{.0141} & 202.63\sd{1.43} \\
\addlinespace[1pt]
\hline
\multirow{2}{*}{TS-GRU}
  & $R^{2}\!\!\uparrow$ & .758\sd{.0006} & .437\sd{.0065} & .735\sd{.0002} & .451\sd{.0029} \\
  & MAE$\downarrow$     & .074\sd{.0002} & .019\sd{.0002} & 2.25\sd{.0020} & 197.42\sd{.411} \\
\addlinespace[1pt]
\multirow{2}{*}{TS-TCN}
  & $R^{2}\!\!\uparrow$ & .692\sd{.0000} & .689\sd{.0061} & .743\sd{.0041} & .459\sd{.0002} \\
  & MAE$\downarrow$     & .086\sd{.0000} & .014\sd{.0002} & 2.27\sd{.0076} & 204.46\sd{.900} \\
\addlinespace[1pt]
\multirow{2}{*}{TS-Former}
  & $R^{2}\!\!\uparrow$ & .747\sd{.0028} & \textbf{.744}\sd{.0098} & .723\sd{.0051} & .465\sd{.0020} \\
  & MAE$\downarrow$     & .076\sd{.0007} & \textbf{.011}\sd{.0003} & 2.29\sd{.0199} & 206.41\sd{1.35} \\
\addlinespace[1pt]
\hline
\multirow{2}{*}{SpikF}
  & $R^{2}\!\!\uparrow$ & \underline{.766}\sd{.0003} & .541\sd{.0019} & .740\sd{.0008} & \underline{.477}\sd{.0047} \\
  & MAE$\downarrow$     & \underline{.072}\sd{.0001} & .016\sd{.0001} & 2.21\sd{.0014} & \underline{186.58}\sd{.700} \\
\hline
\hline
\rowcolor{blue!9}
{\textbf{SpikF-GO}}
  & $R^{2}\!\!\uparrow$ & .762\sd{.0160} & .651\sd{.0032} & \underline{.762}\sd{.0017} & \textbf{.480}\sd{.0097} \\
\rowcolor{blue!9}
  & MAE$\downarrow$     & .077\sd{.0037} & .013\sd{.0000} & \underline{2.17}\sd{.0081} & 189.22\sd{3.28} \\
\addlinespace[1pt]
\rowcolor{blue!9}
{\textbf{SpikF-GO w/ CPG}}
  & $R^{2}\!\!\uparrow$  & \textbf{.769}\sd{.0010} & .669\sd{.0085} & \textbf{.766}\sd{.0018} & .466\sd{.0154} \\
\rowcolor{blue!9}
  & MAE$\downarrow$      & \underline{.072}\sd{.0004} & \underline{.012}\sd{.0002} & \textbf{2.16}\sd{.0116} & 191.21\sd{3.56} \\
\bottomrule
\end{tabular}
\end{table}

\phantomsection
\subsection{Results Across Forecast Horizons}\label{app:C2}

Tables~\ref{tab:app-covid19}--\ref{tab:app-electricity} report forecasting performance across forecast horizons $\{6,12,24,48\}$. Results are averaged over 3 random seeds and reported as mean $\pm$ standard deviation, except for the horizon-12 setting in the main paper, which is averaged over 5 random seeds. Avg.\ denotes the mean score over the four forecast horizons. The compared methods include the top four baseline models for each dataset and our two proposed models.

\begin{table}[t]
\centering
\caption{COVID-19  results across various forecast horizons. Avg.\ averages the mean scores over horizons \{6, 12, 24, 48\}; standard deviations are shown only for individual horizons. \textbf{Bold}/\underline{underline} denote best/second-best based on the unrounded mean values.}
\label{tab:app-covid19}
\begin{tabular}{c:c:cccc:c}
\hline
\textbf{Models} & \textbf{Metric} & $\mathbf{6}$ & $\mathbf{12}$ & $\mathbf{24}$ & $\mathbf{48}$ & \textbf{Avg.} \\
\midrule
\multirow{2}{*}{FourierGNN}
  & $R^{2}\!\!\uparrow$ & \underline{.990}\sd{.002} & .950\sd{.006} & .764\sd{.006} & .408\sd{.002} & .778 \\
  & MAE$\downarrow$ & 16.56\sd{1.3} & 31.50\sd{1.6} & 62.23\sd{1.1} & 103.55\sd{.4} & 53.5 \\
  \addlinespace[1pt]
\hline

\multirow{2}{*}{iSpikformer}
  & $R^{2}\!\!\uparrow$ & .947\sd{.017} & .874\sd{.024} & .689\sd{.021} & .393\sd{.004} & .726 \\
  & MAE$\downarrow$ & 34.04\sd{5.9} & 49.79\sd{5.2} & 73.47\sd{3.0} & 106.38\sd{.6} & 65.9 \\
  \addlinespace[1pt]
\hline
\multirow{2}{*}{TS-GRU}
  & $R^{2}\!\!\uparrow$ & .920\sd{.007} & .842\sd{.006} & .651\sd{.002} & .382\sd{.001} & .699 \\
  & MAE$\downarrow$ & 42.72\sd{1.6} & 56.81\sd{.92} & 79.24\sd{.26} & 108.65\sd{.2} & 71.9 \\
  \addlinespace[1pt]
\hline
\multirow{2}{*}{SpikF}
  & $R^{2}\!\!\uparrow$ & \textbf{.991}\sd{.004} & .961\sd{.006} & .782\sd{.012} & .408\sd{.002} & .786 \\
  & MAE$\downarrow$ & 15.68\sd{2.3} & 28.16\sd{1.8} & 59.83\sd{1.6} & 103.58\sd{.3} & 51.8 \\
  \addlinespace[1pt]
\hline
\rowcolor{blue!9}
{\textbf{SpikF-GO}}
  & $R^{2}\!\!\uparrow$ & \textbf{.991}\sd{.004} & \textbf{.967}\sd{.004} & \textbf{.794}\sd{.018} & \underline{.410}\sd{.002} & \underline{.791} \\
\rowcolor{blue!9}
  & MAE$\downarrow$ & \underline{15.58}\sd{2.6} & \textbf{25.15}\sd{1.7} & \textbf{56.45}\sd{3.0} & \underline{103.18}\sd{.5} & \textbf{50.1} \\
\addlinespace[1pt]
\rowcolor{blue!9}
{\textbf{SpikF-GO w/ CPG}}
  & $R^{2}\!\!\uparrow$ & \textbf{.991}\sd{.003} & \underline{.965}\sd{.002} & \underline{.788}\sd{.021} & \textbf{.420}\sd{.006} & \textbf{.791} \\
\rowcolor{blue!9}
  & MAE$\downarrow$ & \textbf{15.51}\sd{1.6} & \underline{26.09}\sd{.70} & \underline{58.44}\sd{3.4} & \textbf{101.29}\sd{1.} & \underline{50.3} \\
\bottomrule
\end{tabular}
\end{table}

\begin{table}[ht]
\centering
\caption{PEMS-BAY results across various forecast horizons. Avg.\ averages the mean scores over horizons \{6, 12, 24, 48\}; standard deviations are shown only for individual horizons. \textbf{Bold}/\underline{underline} denote best/second-best based on the unrounded mean values.}
\label{tab:app-pemsbay}
\begin{tabular}{c:c:cccc:c}
\hline
\textbf{Models} & \textbf{Metric} & $\mathbf{6}$ & $\mathbf{12}$ & $\mathbf{24}$ & $\mathbf{48}$ & \textbf{Avg.} \\
\midrule
\multirow{2}{*}{FourierGNN}
  & $R^{2}\!\!\uparrow$ & .847\sd{.001} & .733\sd{.003} & .508\sd{.001} & .123\sd{.008} & .553 \\
  & MAE$\downarrow$ & \textbf{1.62}\sd{.002} & \textbf{2.16}\sd{.002} & 3.04\sd{.002} & 4.39\sd{.010} & 2.80 \\
\hline
\addlinespace[1pt]
\multirow{2}{*}{SpikeTCN}
  & $R^{2}\!\!\uparrow$ & .857\sd{.000} & .747\sd{.001} & .539\sd{.001} & .078\sd{.012} & .555 \\
  & MAE$\downarrow$ & 1.72\sd{.000} & 2.25\sd{.002} & 3.09\sd{.008} & 4.48\sd{.013} & 2.89 \\
  \addlinespace[1pt]
\hline
\multirow{2}{*}{TS-TCN}
  & $R^{2}\!\!\uparrow$ & .854\sd{.003} & .743\sd{.004} & .520\sd{.013} & .189\sd{.000} & .577 \\
  & MAE$\downarrow$ & 1.74\sd{.012} & 2.27\sd{.008} & 3.13\sd{.014} & 4.46\sd{.000} & 2.90 \\
  \addlinespace[1pt]
\hline
\multirow{2}{*}{SpikF}
  & $R^{2}\!\!\uparrow$ & .862\sd{.001} & .740\sd{.001} & .504\sd{.000} & .120\sd{.000} & .557 \\
  & MAE$\downarrow$ & \underline{1.66}\sd{.001} & 2.21\sd{.001} & 3.10\sd{.001} & 4.41\sd{.001} & 2.85 \\
  \addlinespace[1pt]
\hline
\rowcolor{blue!9}
{\textbf{SpikF-GO}}
  & $R^{2}\!\!\uparrow$ & \underline{.870}\sd{.001} & \underline{.762}\sd{.002} & \underline{.545}\sd{.004} & \underline{.209}\sd{.005} & \underline{.597} \\
\rowcolor{blue!9}
  & MAE$\downarrow$ & \textbf{1.62}\sd{.000} & \underline{2.17}\sd{.008} & \underline{3.03}\sd{.002} & \textbf{4.33}\sd{.010} & \underline{2.79} \\
\addlinespace[1pt]
\rowcolor{blue!9}
{\textbf{SpikF-GO w/ CPG}}
  & $R^{2}\!\!\uparrow$ & \textbf{.873}\sd{.000} & \textbf{.766}\sd{.002} & \textbf{.569}\sd{.000} & \textbf{.225}\sd{.006} & \textbf{.608} \\
\rowcolor{blue!9}
  & MAE$\downarrow$ & \textbf{1.62}\sd{.003} & \textbf{2.16}\sd{.012} & \textbf{3.02}\sd{.023} & \underline{4.34}\sd{.028} & \textbf{2.79} \\
\bottomrule
\end{tabular}
\end{table}

\begin{table}[ht]
\centering
\caption{Solar results across various forecast horizons. Avg.\ averages the mean scores over horizons \{6, 12, 24, 48\}; standard deviations are shown only for individual horizons. \textbf{Bold}/\underline{underline} denote best/second-best based on the unrounded mean values.}
\label{tab:app-solar}
\begin{tabular}{c:c:cccc:c}
\hline
\textbf{Models} & \textbf{Metric} & $\mathbf{6}$ & $\mathbf{12}$ & $\mathbf{24}$ & $\mathbf{48}$ & \textbf{Avg.} \\
\midrule
\multirow{2}{*}{FourierGNN}
  & $R^{2}\!\!\uparrow$ & \textbf{.767}\sd{.001} & \textbf{.742}\sd{.001} & \underline{.703}\sd{.001} & \textbf{.670}\sd{.001} & \textbf{.721} \\
  & MAE$\downarrow$ & \textbf{8.56}\sd{.008} & \textbf{9.03}\sd{.019} & \underline{9.80}\sd{.020} & \underline{10.47}\sd{.02} & \underline{9.47} \\
\hline
\addlinespace[1pt]
\multirow{2}{*}{SpikeTCN}
  & $R^{2}\!\!\uparrow$ & .743\sd{.007} & .706\sd{.006} & .656\sd{.005} & .620\sd{.007} & .681 \\
  & MAE$\downarrow$ & 9.26\sd{.148} & 9.84\sd{.098} & 10.74\sd{.12} & 11.29\sd{.09} & 10.28 \\
  \addlinespace[1pt]
\hline
\multirow{2}{*}{TS-TCN}
  & $R^{2}\!\!\uparrow$ & .740\sd{.004} & .707\sd{.007} & .658\sd{.004} & .621\sd{.006} & .682 \\
  & MAE$\downarrow$ & 9.28\sd{.089} & 9.80\sd{.100} & 10.67\sd{.11} & 11.30\sd{.06} & 10.26 \\
  \addlinespace[1pt]
\hline
\multirow{2}{*}{SpikF}
  & $R^{2}\!\!\uparrow$ & .735\sd{.000} & .712\sd{.001} & .667\sd{.002} & .633\sd{.001} & .687 \\
  & MAE$\downarrow$ & 9.20\sd{.022} & 9.61\sd{.032} & 10.43\sd{.045} & 11.14\sd{.024} & 10.10 \\
  \addlinespace[1pt]
\hline
\rowcolor{blue!9}
{\textbf{SpikF-GO}}
  & $R^{2}\!\!\uparrow$ & \underline{.765}\sd{.004} & \underline{.740}\sd{.005} & .702\sd{.008} & \underline{.665}\sd{.014} & .718 \\
\rowcolor{blue!9}
  & MAE$\downarrow$ & 8.61\sd{.068} & \underline{9.18}\sd{.065} & 9.87\sd{.098} & 10.51\sd{.164} & 9.54 \\
\addlinespace[1pt]
\rowcolor{blue!9}
{\textbf{SpikF-GO w/ CPG}}
  & $R^{2}\!\!\uparrow$ & \textbf{.767}\sd{.011} & \textbf{.742}\sd{.005} & \textbf{.707}\sd{.016} & \underline{.665}\sd{.028} & \underline{.720} \\
\rowcolor{blue!9}
  & MAE$\downarrow$ & \underline{8.60}\sd{.144} & \textbf{9.03}\sd{.106} & \textbf{9.66}\sd{.203} & \textbf{10.40}\sd{.37} & \textbf{9.42} \\
\bottomrule
\end{tabular}
\end{table}

\begin{table}[t]
\centering
\caption{Electricity results across various forecast horizons. Avg.\ averages the mean scores over horizons \{6, 12, 24, 48\}; standard deviations are shown only for individual horizons. \textbf{Bold}/\underline{underline} denote best/second-best based on the unrounded mean values.}
\label{tab:app-electricity}
\begin{tabular}{c:c:cccc:c}
\hline
\textbf{Models} & \textbf{Metric} & $\mathbf{6}$ & $\mathbf{12}$ & $\mathbf{24}$ & $\mathbf{48}$ & \textbf{Avg.} \\
\midrule
\multirow{2}{*}{FourierGNN}
  & $R^{2}\!\!\uparrow$ & .959\sd{.000} & .912\sd{.001} & .799\sd{.000} & .598\sd{.002} & .817 \\
  & MAE$\downarrow$ & 65.94\sd{.20} & 97.13\sd{.47} & 157.46\sd{.1} & 241.78\sd{.0} & 141 \\
\hline
\addlinespace[1pt]
\multirow{2}{*}{Spikformer w/ CPG}
  & $R^{2}\!\!\uparrow$ & .963\sd{.003} & \underline{.930}\sd{.006} & \underline{.858}\sd{.006} & \textbf{.761}\sd{.011} & \underline{.878} \\
  & MAE$\downarrow$ & 65.59\sd{1.1} & \textbf{84.65}\sd{.43} & \textbf{118.99}\sd{2.} & \textbf{168.92}\sd{.8} & \textbf{110} \\
\hline
\addlinespace[1pt]
\multirow{2}{*}{TS-Former}
  & $R^{2}\!\!\uparrow$ & .961\sd{.005} & .925\sd{.009} & .835\sd{.006} & .738\sd{.012} & .865 \\
  & MAE$\downarrow$ & 64.91\sd{1.1} & \underline{84.78}\sd{.62} & \underline{121.25}\sd{.4} & \underline{170.19}\sd{3.} & \underline{110} \\
  \addlinespace[1pt]
\hline
\multirow{2}{*}{SpikF}
  & $R^{2}\!\!\uparrow$ & .958\sd{.000} & .910\sd{.001} & .795\sd{.002} & .596\sd{.000} & .815 \\
  & MAE$\downarrow$ & 69.35\sd{.08} & 99.99\sd{.12} & 157.91\sd{.0} & 238.06\sd{.1} & 141 \\
  \addlinespace[1pt]
 \hline
\rowcolor{blue!9}
{\textbf{SpikF-GO}}
  & $R^{2}\!\!\uparrow$ & \underline{.967}\sd{.001} & .927\sd{.007} & .834\sd{.005} & .718\sd{.000} & .862 \\
\rowcolor{blue!9}
  & MAE$\downarrow$ & \textbf{63.25}\sd{.19} & 91.15\sd{1.3} & 144.60\sd{2.} & 213.32\sd{.9} & 128 \\
\addlinespace[1pt]
\rowcolor{blue!9}
{\textbf{SpikF-GO w/ CPG}}
  & $R^{2}\!\!\uparrow$ & \textbf{.971}\sd{.001} & \textbf{.942}\sd{.003} & \textbf{.859}\sd{.003} & \underline{.745}\sd{.001} & \textbf{.879} \\
\rowcolor{blue!9}
  & MAE$\downarrow$ & \underline{64.29}\sd{.28} & 89.14\sd{1.1} & 142.01\sd{.1} & 211.64\sd{1.} & 127 \\
\bottomrule
\end{tabular}
\end{table}

\end{document}